\documentclass[10pt,twocolumn,letterpaper]{article}

\usepackage[pagenumbers]{iccv} % To force page numbers, e.g. for an arXiv version
\graphicspath{{./pics/}{../}}

\definecolor{iccvblue}{rgb}{0.21,0.49,0.74}
\usepackage[pagebackref,breaklinks,colorlinks,allcolors=iccvblue]{hyperref}

\usepackage{tcolorbox}
\usepackage{multirow}
\usepackage{soul}
\usepackage{graphicx}
\usepackage{float}
\usepackage{colortbl}

\usepackage{fontawesome5}

\definecolor{mycolor}{RGB}{142, 85, 115} % light purple, a little shade

\definecolor{mygreen}{RGB}{255,127,80}

\newcommand{\minisection}[1]{\vspace{.03in}\noindent{\textbf{#1}.}}
\def\paperID{5870} % *** Enter the Paper ID here
\def\confName{ICCV}
\def\confYear{2025}

\title{Seg-\textcolor{mycolor}{Zero}: Reasoning-Chain Guided  Segmentation via Cognitive Reinforcement}
\author{
Yuqi Liu$^{1*}$ \hspace{1pt}
Bohao Peng$^{1*}$ \hspace{1pt}
Zhisheng Zhong$^{1*}$ \hspace{1pt}
Zihao Yue$^{3}$ \hspace{1pt}
Fanbin Lu$^{1}$ \hspace{1pt}
Bei Yu$^{1}$ \hspace{1pt}
Jiaya Jia$^{2}$ \hspace{1pt}
\\ 
{\small The Chinese University of Hong Kong$^{1}$ \hspace{8pt} The Hong Kong University of Science and Technology$^{2}$ } \\
{\small Renmin University of China $^{3}$   \hspace{8pt} $^*$Equal \hspace{8pt} \href{https://github.com/JIA-Lab-research/Seg-Zero}{Code \faGithub}} 
}

\begin{document}

\twocolumn[{%
    \renewcommand\twocolumn[1][]{#1}%
    \maketitle
    \vspace{-18pt}
    \begin{center}
        \captionsetup{type=figure}
        \centering
        \includegraphics[width=1.0\linewidth]{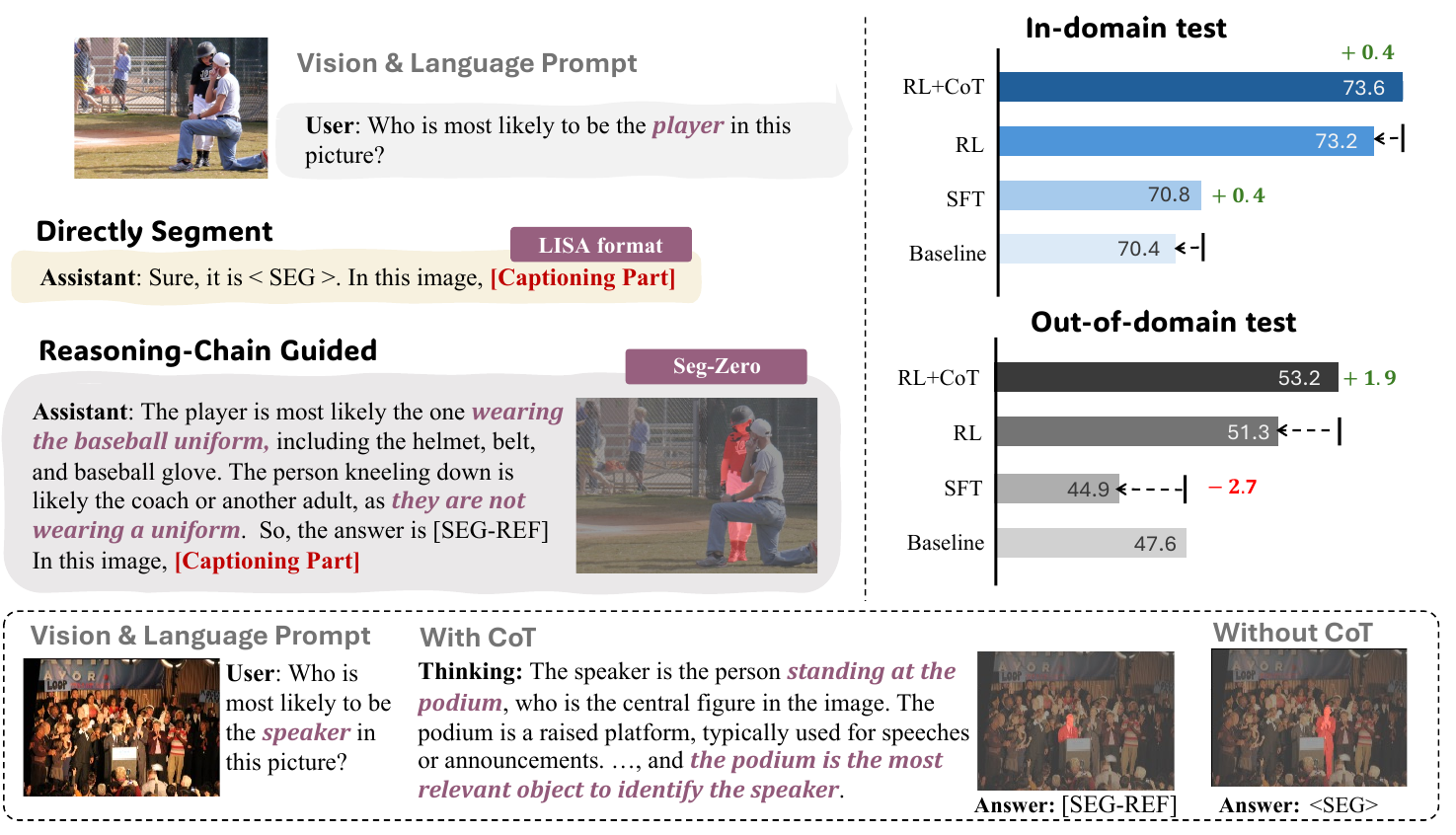}
        \captionof{figure}{Seg-\textcolor{mycolor}{Zero} generates a reasoning chain before producing the final segmentation mask. It utilizes a pure reinforcement learning (RL) strategy, learning the reasoning process from \textcolor{mycolor}{zero}. In comparison to supervised fine-tuning (SFT), the RL-based model demonstrates superior performance on both in-domain and out-of-domain data, and the integration of reasoning chain further enhances its effectiveness.}
        \label{fig:teaser_figure}
    \end{center}%
    
}]

\begin{abstract}
Traditional methods for reasoning segmentation rely on supervised fine-tuning with categorical labels and simple descriptions, limiting its out-of-domain generalization and lacking explicit reasoning processes. To address these limitations, we propose Seg-Zero, a novel framework that demonstrates remarkable generalizability and derives explicit chain-of-thought reasoning through cognitive reinforcement. Seg-Zero introduces a decoupled architecture consisting of a reasoning model and a segmentation model. The reasoning model interprets user intentions, generates explicit reasoning chains, and produces positional prompts, which are subsequently used by the segmentation model to generate precious pixel-level masks. 
We design a sophisticated reward mechanism that integrates both format and accuracy rewards to effectively guide optimization directions. Trained exclusively via reinforcement learning with GRPO and without explicit reasoning data, Seg-Zero achieves robust zero-shot generalization and exhibits emergent test-time reasoning capabilities. 
Experiments show that Seg-Zero-7B achieves a zero-shot performance of 57.5 on the ReasonSeg benchmark, surpassing the prior LISA-7B by 18\%. This significant improvement highlights Seg-Zero's ability to generalize across domains while presenting an explicit reasoning process.  
\end{abstract}
    
\section{Introduction}
\label{sec:introduction}

Reasoning segmentation generates pixel-wise masks by interpreting implicit queries through logical reasoning. This task shows significant potential in real-world applications, such as robots. Unlike conventional segmentation tasks that rely on simple categorical labels (e.g., ``person'' or ``car''), reasoning segmentation addresses more complex and nuanced queries, such as ``identify food that provides sustained energy.'' Such queries require logical reasoning and the integration of cross-domain knowledge to produce accurate segmentation masks.

Early attempts \cite{lai2024lisa,bai2025onetokensegall,ren2024pixellm}, such as LISA \cite{lai2024lisa}, have explored the use of multimodal large language models (MLLMs) to enhance reasoning segmentation capabilities, These methods bridge the gap between MLLMs and segmentation models by leveraging implicit semantic tokens.
However, typical methods \cite{lai2024lisa,ren2024pixellm,chen2024sam4mllm} rely solely on supervised fine-tuning (SFT) applied to mixed datasets containing only simple categorical information or basic factual descriptions \cite{yu2016refcoco,kazemzadeh2014referitgame,he2022partimagenet}. Although this paradigm effectively aligns MLLMs \cite{liu2023llava,liu2024llava15,wang2024qwen2vl} with segmentation models \cite{kirillov2023sam} in specific datasets, we observe that it lacks generalization capabilities. This can be demonstrated by: (i) Although existing methods excel on in-domain data, their performance significantly degrades on out-of-distribution (OOD) samples. (ii) SFT inevitably leads to catastrophic forgetting of general capabilities. 
(iii) The lack of an explicit reasoning process hinders their effectiveness in complex scenarios.
These limitations motivate us to enhance general segmentation capabilities and improve reasoning performance by integrating an explicit reasoning process. 

Recent studies \cite{guo2025deepseekr1} demonstrate that training with pure reinforcement learning (RL) activates the emergent test-time reasoning process, highlighting that reward-driven optimization is effective in enhancing model reasoning ability. Moreover, this approach often promotes generalization rather than overfitting to specific datasets.
Inspired by this, we introduce Seg-\textcolor{mycolor}{Zero}, a novel framework designed to enhance reasoning and cognitive capabilities for reasoning segmentation. 
Seg-Zero adopts a decoupled architecture, including a reasoning model and a segmentation model. The reasoning model is an MLLM capable of processing both image and user instructions. It outputs not only region-level bounding boxes (bbox) but also pixel-level points to precisely localize the target object.
Subsequently, the segmentation model utilizes the bbox and points to produce pixel-level segmentation masks. 
% This design eliminates the need for auxiliary alignment between MLLMs and segmentation models, which enhances flexibility and easy to integrate state-of-the-art (SOTA) models for each component.

% In this section, we
%  explore the potential of LLMs to develop reasoning capabilities without any supervised data,
%  focusing on their self-evolution through a pure reinforcement learning process
During training, we employ pure reinforcement learning, specifically GRPO~\cite{shao2024deepseekmath}, to fine-tune the reasoning model while keeping the segmentation model frozen. 
Rather than constructing datasets with explicitly annotated reasoning processes, we investigate the self-evolution potential of MLLM to develop reasoning capabilities, thereby achieving emergent reasoning from \textcolor{mycolor}{zero}.
%Instead of curating datasets that explicitly include reasoning processes, we explore the potential of self-evolution of MLLMs to develop reasoning capabilities, enabling the model to emerge reasoning from \textcolor{mycolor}{zero}.
To achieve this, we develop a sophisticated reward mechanism to enhance the reasoning process and regulate the output. 
%This mechanism is designed to strengthen the test-time reasoning process and produce accurate location information, thereby improving the handling of complex segmentation instructions. 
These reward functions comprise two types: format rewards, which enforce constraints on the structure of the reasoning process and segmentation outputs, and accuracy rewards, which are calculated based on intersection over union (IoU) and L1 distance metrics. 
As illustrated in \Cref{fig:teaser_figure}, by leveraging optimized reward-driven reinforcement learning, our Seg-Zero exhibits emergent test-time reasoning abilities, similar to those demonstrated in LLMs~\cite{guo2025deepseekr1,openaio1}. This reasoning process enables the model to effectively handle complex instructions by breaking them down into sequential analytical steps, thus achieving the precise localization of target objects. Seg-Zero demonstrates exceptional performance on both in-domain and OOD data, significantly exceeding the model trained through SFT. Furthermore, Seg-Zero maintains robust visual QA capability, without the need for VQA training data.

Experimental results show that, with only 9,000 training samples derived from RefCOCOg \cite{yu2016refcoco}, our Seg-Zero-7B exhibits strong test-time reasoning capabilities and achieves superior generalization performance compared to models of the same scale. It achieves a \textcolor{mycolor}{zero}-shot performance of 57.5 on ReasonSeg \cite{lai2024lisa}, surpassing the previous LISA-7B by 18\%. 

We summarize our contributions as follows:
\begin{itemize}
    \item We propose Seg-Zero, a novel architecture designed for reasoning segmentation. Through the pure RL algorithm, Seg-Zero exhibits emergent reasoning abilities.
    \item We present a detailed comparison between SFT and RL, as well as the integration of reasoning chain. Results demonstrates that RL, combined with the reasoning chain, consistently enhances model performance.
    \item Extensive experiments demonstrate the effectiveness of our design and offer valuable insight for fine-tuning models using RL.
\end{itemize}

\section{Related Works}
\label{sec:relatedworks}

\subsection{Reasoning in Large Models}
In recent years, Large Language Models (LLMs) have exhibited remarkable reasoning capabilities. By extending the length of the Chain-of-Thought (CoT) reasoning process, OpenAI-o1 \cite{openaio1} introduces inference-time scaling, significantly improving its reasoning performance. In the research community, several studies have attempted to achieve test-time scaling through various approaches, including process-based reward models \cite{wang2023mathshepherd,uesato2022solvingmath,lightman2023stepbystep}, reinforcement learning (RL) \cite{kumar2024training,shao2024deepseekmath}, and search algorithms \cite{feng2023alphazerolike,trinh2024solving}. In particular, the recent DeepSeek-R1 \cite{guo2025deepseekr1}, which uses the GRPO \cite{shao2024deepseekmath} algorithm, achieves superior performance with only a few thousand RL training steps.
Building on advances in the LLMs community, several recent works have attempted to leverage the reasoning capabilities of MLLMs \cite{openr1multimodal,r1v}. For example, Open-R1-Multimodal \cite{openr1multimodal} emphasizes mathematical reasoning, while R1-V \cite{r1v} shows exceptional performance in counting tasks. However, these works primarily address high-level reasoning and do not consider fine-grained pixel-level understanding of images. To fill this gap, our Seg-Zero is designed to enhance pixel-level reasoning through reinforcement learning.

\subsection{Semantic Segmentation with Reasoning}
Semantic segmentation aims at predicting segmentation masks for specific classes. Numerous studies~\cite{ronneberger2015u,long2015fully, lin2017refinenet,chen2017deeplab, chen2017rethinking, zhao2017pyramid,badrinarayanan2017segnet, cheng2021per}, including DeepLab~\cite{chen2018encoder}, MaskFormer~\cite{cheng2022masked} and SAM~\cite{kirillov2023sam} have made significant progress in this task, making it a well-addressed problem. 
Instead of segmenting objects with explicit class labels, referring expression segmentation \cite{kazemzadeh2014referitgame,yu2016refcoco} focuses on segmenting target objects based on short, explicit text queries. %This task is more challenging as images often contain multiple instances of the same object class with varying attributes, requiring the model to identify and segment the instance that most closely matches the text query. 
LISA \cite{lai2024lisa} advances this field further by introducing the reasoning segmentation task. In this task, text queries are either more intricate or longer, demanding models with strong reasoning capabilities to accurately interpret and segment the target objects.

\subsection{MLLMs for Segmentation}
Since LISA \cite{lai2024lisa,yang2023lisa++} introduced the ` \textless SEG \textgreater ' token to bridge the gap between MLLMs and segmentation models, several subsequent works \cite{chen2024sam4mllm,ren2024pixellm,bai2025onetokensegall} have explored the use of MLLMs for segmentation tasks. Most of these approaches, including OneTokenSegAll \cite{bai2025onetokensegall} and PixelLM \cite{ren2024pixellm}, follow LISA's paradigm by using special tokens to connect MLLMs with segmentation models. However, this design necessitates extensive data to fine-tune both the MLLM and the segmentation decoder, and may even compromise the pixel precious of the original segmentation models. 
% In contrast, SAM4MLLM \cite{chen2024sam4mllm} adopts a decoupled architecture, separating the MLLM and the segmentation model, but it does not fully leverage the reasoning capabilities of MLLMs, limiting its overall performance. 
Our proposed Seg-Zero also employs a decoupled design for ease of adoption, while further leveraging the reasoning ability of MLLMs to achieve superior results.

\begin{figure}[t]
  \centering
   \includegraphics[width=1.\linewidth]{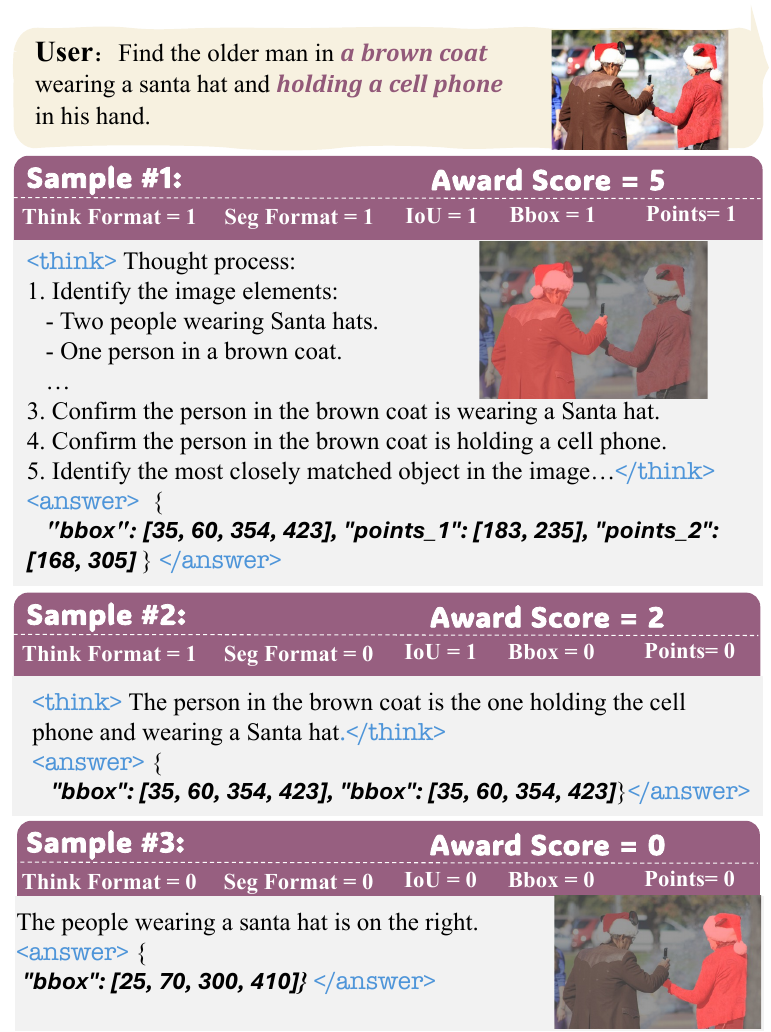}
   \caption{Illustration of our RL training process. In this case, the model generates three samples by itself, calculates the rewards, and optimizes towards samples that achieve higher rewards.}
   \label{fig:gpro_illustration}
\end{figure}

\begin{figure*}[t]
  \centering
   \includegraphics[width=.96\linewidth]{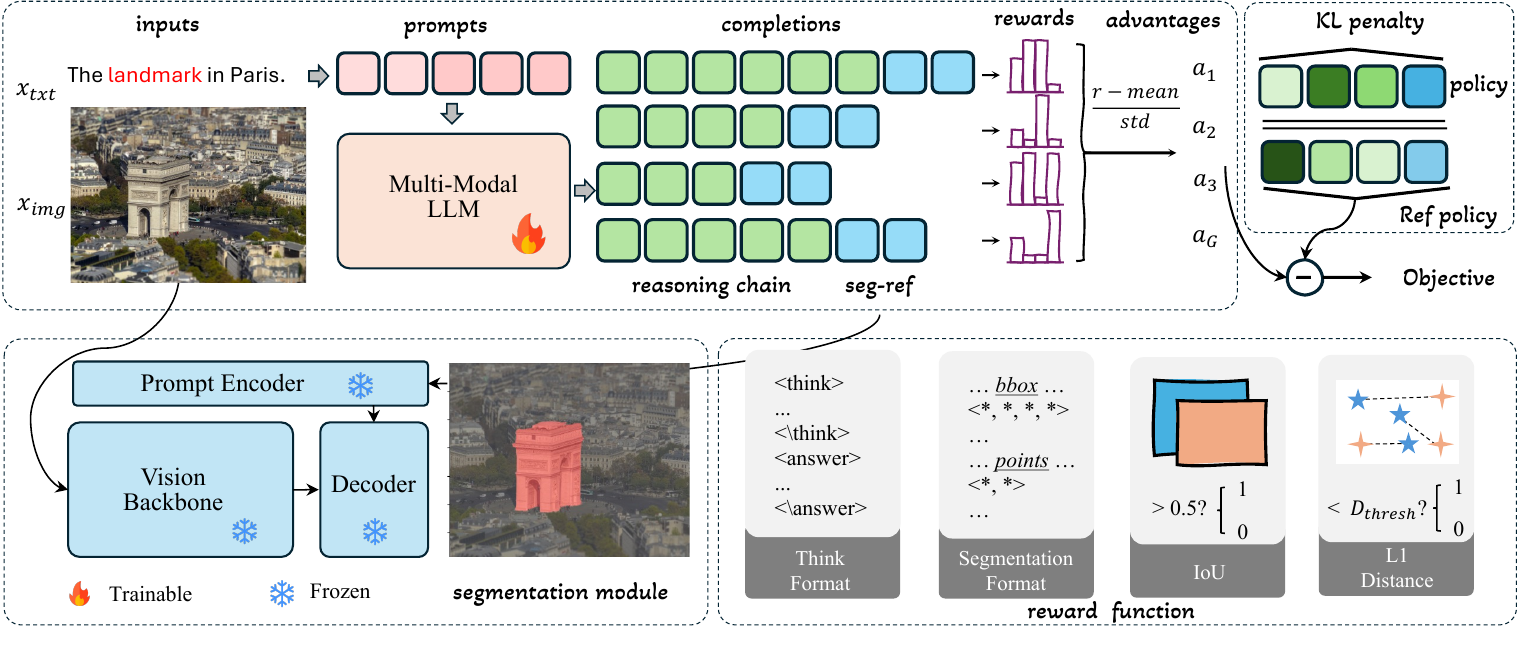}
   \caption{Seg-Zero includes a reasoning model and a segmentation model. The reasoning model is a MLLM that generates a reasoning chain and provides segmentation prompts. Subsequently the segmentation model produces pixel-wise mask.}
   \label{fig:architecture}
\end{figure*}

\section{Method}

In this section, we introduce our Seg-Zero model and the associated reinforcement learning framework. 
We first describe how we address the segmentation problem in \Cref{sec:problem}. Next, we present the architecture of the Seg-Zero in \Cref{sec:model}. Finally, we describe the reward functions (\Cref{sec:rewards}) and the training details (\Cref{sec:training}) in the reinforcement learning framework.

\subsection{Pipeline Formulation}
\label{sec:problem}

Given an image $\mathbf{I}$ and a label $\mathbf{T}$ , the segmentation task aims to produce a binary segmentation mask $\mathbf{M}$ that accurately identifies the region corresponding to $\mathbf{T}$. The label $\mathbf{T}$ can vary in complexity, ranging from a simple class label (e.g., ``bird''), to a straightforward phrase (e.g., ``woman in blue''), or even to long and intricate expressions (e.g., ``The unusual thing in the image''). The latter two types of expression require the model to perform reasoning to accurately segment the most relevant objects. 

Inspired by recent advancements in the reasoning capabilities of large models \cite{shao2024deepseekmath,r1v,guo2025deepseekr1}, we leverage this ability to develop a pipeline for reasoning-based segmentation.
Specifically, we decouple the reasoning process and the segmentation process. We first employ reinforcement learning to an MLLM to activate its reasoning ability, enabling it to generate the reasoning process and produce accurate bounding box $\mathbf{B}$ and two points $\mathbf{P_1,P_2}$ that best localize the target object. These bounding box and points are then used as prompts for SOTA segmentation models \cite{kirillov2023sam,ravi2024sam2} to produce fine-grained segmentation masks. Seg-Zero is trained using reinforcement learning, as illustrated in \Cref{fig:gpro_illustration}.
%In contrast to the composite design in LISA \cite{lai2024lisa}, our decoupled design provides clear advantages, as it enables us to easily adopt newly developed SOTA models for each component.
% , and alleviate the resource consumption in RL, keeps segementation model accurarcy.

\subsection{Seg-\textcolor{mycolor}{Zero} Model}
\label{sec:model}

Current MLLMs \cite{wang2024qwen2vl,li2024mgm,zhong2024lyra,bai2025qwen25,liu2024llava15} exhibit impressive performance in processing multi-modal inputs but are unable to generate fine-grained segmentation masks. Conversely, modern segmentation models \cite{kirillov2023sam,ravi2024sam2} provide fine-grained segmentation ability but lack robust reasoning capabilities. To bridge this gap, we propose Seg-Zero, a framework that includes a reasoning model and a segmentation model. Additionally, we introduce the novel strategy to effectively activate the reasoning ability of MLLM within the framework. Its whole architecture is shown in \Cref{fig:architecture}.

\minisection{Reasoning Model}
We employ Qwen2.5-VL \cite{bai2025qwen25} as our reasoning model $\mathcal{F}_{reason}$. Although Qwen2.5-VL demonstrates exceptional performance in object detection by predicting the bbox, this region-level bbox is insufficient to provide more fine-grained pixel-level localization. Unlike object detection, segmentation requires a more precise understanding of pixel-level details, as multiple objects may exist within a single bounding box. Therefore, in addition to the bounding box, we also incorporate points that lie within the target object to improve localization accuracy. 
During the reinforcement learning stage, the format rewards are employed to ensure the model generates structured outputs, which are subsequently processed by a post-processing function $\mathcal{G}$ to extract the bounding box $\mathbf{B}$ and the two points $\mathbf{P_1,P_2}$ . 
This process can be formulated as follows: 

\begin{equation}
    \mathbf{B,P_1,P_2} = \mathcal{G}(\mathcal{F}_{reason}(\mathbf{I,T})).
\end{equation}

\minisection{Segmentation Model}
Modern segmentation models \cite{kirillov2023sam,ravi2024sam2} accept various types of prompt,
including bounding boxes and points, to generate accurate segmentation masks.
We employ SAM2 \cite{ravi2024sam2} as our segmentation model $\mathcal{F}_{seg}$ due to its superior performance and efficient inference speed.
Leveraging the bounding boxes and points provided by the reasoning model, the segmentation model can generate a precise, fine-grained mask for the target object.
This process can be formally expressed as follows:
\begin{equation}
    \mathbf{M} = \mathcal{F}_{seg}(\mathbf{B,P_1,P_2}).
\end{equation}

%%% seperate

% \begin{table*}[t!]
% \centering
% \tcbset{
%   colback=gray!5!white, % Background color
%   colframe=mycolor!60,       % Frame color
%   width=0.92\linewidth,     % Box width
%   boxrule=1pt,          % Border thickness
%   arc=4mm,              % Rounded corners
%   left=5pt,             % Left padding
%   right=5pt,            % Right padding
%   top=5pt,              % Top padding
%   bottom=5pt,           % Bottom padding
% }
% \begin{tcolorbox}[title=User Prompt for Seg-Zero]
% \footnotesize 

% ``Please find `\textcolor{red}{\{Question\}}' with bbox and points." \

% ``Compare the difference between objects and find the most closely matched one." \

% ``Output the thinking process in \texttt{<think>} \texttt{<\textbackslash think>} and final answer in \texttt{<\textbackslash answer>} \texttt{<\textbackslash answer>} tags." \

% ``Output the one bbox and center points of two largest inscribed circles inside the interested object in JSON format." \

% ``i.e., \texttt{<think>} thinking process here \texttt{<\textbackslash think>}" \

% \texttt{<answer>} \{`bbox': [10,100,200,210], `points\_1': [30,110], `points\_2': [35,180]\} \texttt{<\textbackslash answer>}" \

% \normalsize
% \end{tcolorbox}

% \caption{User prompt for Seg-Zero. `\textcolor{red}{\{Question\}}' is replaced with object description $\mathbf{T}$ in the training and inference.}
% \label{tab:user_prompt_template}
% \end{table*}

\begin{figure*}[tb!]
  \centering
   \includegraphics[width=0.95\linewidth]{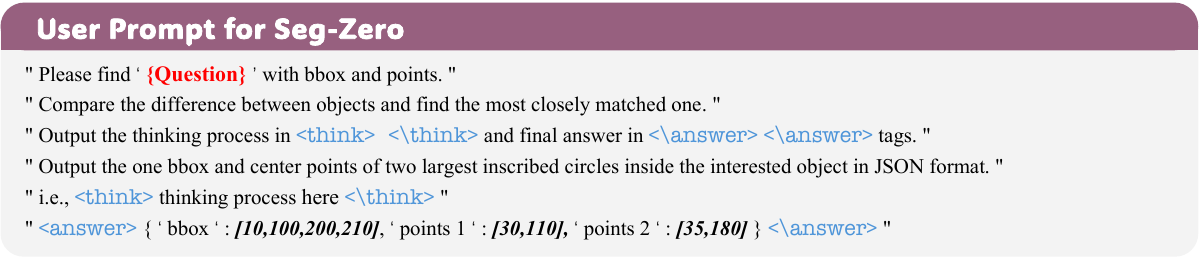}
    \caption{User prompt for Seg-Zero. `\textcolor{red}{\{Question\}}' is replaced with object description $\mathbf{T}$ in the training and inference.}
    \label{tab:user_prompt_template}
\end{figure*}

\minisection{Test-time Reasoning}
Reasoning is the crucial part in reasoning segmentation tasks. Inspired by DeepSeek-R1-Zero, we intentionally avoid using any explicit Chain-of-Thought (CoT) data to teach Seg-Zero reasoning skills. Instead, we aim to activate its reasoning capabilities from \textcolor{mycolor}{zero}, enabling the model to autonomously generate a logical CoT before producing the final answer. To achieve this, we design a structured user prompt and a sophisticated reward mechanism to guide the reasoning model toward the correct optimization direction. As shown in \Cref{tab:user_prompt_template}, the user prompt instructs Seg-Zero to analyze and compare objects in the image, beginning by generating a reasoning process, followed by the final answer in a pre-defined format.
The reward mechanism then evaluates the answers and directs the optimization process, as illustrated in \Cref{fig:gpro_illustration}.

\subsection{Reward Functions}
\label{sec:rewards}

Reward functions play a pivotal role in reinforcement learning, as they determine the optimization directions of the model. We manually design the following five reward functions for reinforcement learning.

\minisection{Thinking Format Reward}
This reward is designed to force the model engage in a structured thinking process. It guides the model output its reasoning steps within the \textless think\textgreater ~and~  \textless /think\textgreater tags, and the final answer is included between the \textless answer\textgreater ~and~ \textless /answer\textgreater tags.

% \textbf{TO BE Done} answer inside the thinking 

\minisection{Segmentation Format Reward}
Different from counting or other QA tasks, the segmentation task is highly dependent on the format of the answer. We provide two types of segmentation format rewards: soft and strict. Under soft constraints, if the keywords \textit{bbox} and \textit{points} appear in the answer, and their corresponding values consist of four and two coordinates, respectively, the format is considered correct. Under strict constraints, the format is only considered correct if the model outputs exact keywords (e.g., \textit{bbox}, \textit{points\_1}, \textit{points\_2}) in the required structure.
% However, strict constrains is not ideal, as it slows down the convergence speed during training when the model struggles to generate entirely consistent keywords (e.g., bbox) or even becomes "stuck". To address this issue, we propose a more flexible reward mechanism for answer formatting. Specifically, if the terms \textit{bbox} and \textit{points} appear in the answer, and their corresponding values consist of four and two coordinates, respectively, the format is considered correct. 
% an ablation can be done for this claim.

\minisection{Bbox IoU Reward}
This reward evaluates the IoU between the predicted bbox and the ground-truth bbox. A reward of 1 is assigned if their IoU greater than 0.5; otherwise, the reward is 0.

\minisection{Bbox L1 Reward}
This reward evaluates the L1 distance between the predicted bbox and the ground-truth bbox. A reward of 1 is assigned if their L1 distance less than 10 pixels; otherwise, the reward is 0.

\minisection{Point L1 Reward}
This reward evaluates the L1 distance between the predicted points and the ground-truth points.
We first determine whether the predicted points are inside the bounding box. Then the reward is set to 1 if the minimal distance between the predicted points and the ground-truth points is less than 100 pixels; otherwise, the reward is 0.

\subsection{Training}
\label{sec:training}

We build the training data from publicly available segmentation datasets and train our Seg-Zero using the GRPO algorithm.

\minisection{Data Preparation}
The training data is generated using the original mask annotations from existing referring expression segmentation datasets (e.g., RefCOCOg \cite{yu2016refcoco}). Based on the mask, we extract the leftmost, topmost, rightmost, and bottommost pixels of the mask to generate the bounding box $\mathbf{B}$. Additionally, we compute the center points of the two largest inscribe circles within the mask, denoted as $\mathbf{P_1}$ and $\mathbf{P_2}$ . Consequently, the ground truth data comprises the bbox coordinates [$\mathbf{B}_{x1}$, $\mathbf{B}_{y1}$, $\mathbf{B}_{x2}$, $\mathbf{B}_{y2}$] and the coordinates of the two center points [$\mathbf{P_1}_{x}$, $\mathbf{P_1}_{y}$] and [$\mathbf{P_2}_x$, $\mathbf{P_2}_y$]. We do not incorporate any CoT processing into the training data. To ensure consistency, all images are rescaled to a uniform resolution of 840x840 pixels.

\minisection{GRPO}
We do not include any \textit{reasoning data} for a cold-start training process to teach the model's reasoning ability.
Instead, we let our Seg-Zero evolve from \textcolor{mycolor}{zero}. Specifically, we initiate training directly from the pre-trained Qwen2.5-VL-3B model, utilizing the aforementioned rewards and applying the GRPO algorithm \cite{shao2024deepseekmath}. We illustrate our RL training process in \Cref{fig:gpro_illustration}.
% \textcolor{red}{some calculate details or ignore}

\section{Experiment}

\subsection{Experimental Settings}
\textbf{Datasets.} We training our Seg-Zero with only 9,000 samples adopted from RefCOCOg, using the data preparation strategy mentioned in \Cref{sec:training}. The test data includes ReasonSeg \cite{lai2024lisa} and RefCOCO(+/g) \cite{yu2016refcoco}.
% RefCOCO consists of 5,657 samples in  val(3811) testA(1975) and 5,095 samples in testB(1810). RefCOCO+ contains 3,805 (val) 5,726 samples in testA(1975) and 4,889 in testB(1798), while RefCOCOg has 2,573 validation samples and 5,023 test samples. Additionally, ReasonSeg includes 779 test samples.

% on an 8xH200 GPU server
\minisection{Implementation Details}
We employ Qwen2.5-VL-3B \cite{bai2025qwen25} and SAM2-Large \cite{ravi2024sam2}  as our default reasoning model and segmentation model, respectively. Seg-Zero is trained using the DeepSpeed \cite{rasley2020deepspeed} library. During training, we use a total batch size of 16 with a sampling number of 8 per training step. The initial learning rate is set to 1e-6 and the weight decay is 0.01.

\minisection{Evaluation Metrics}
Following previous works \cite{yu2016refcoco,kazemzadeh2014referitgame}, we calculate gIoU and cIoU. The gIoU is the average of all per-image Intersection-over-Unions (IoUs), while the cIoU calculates the cumulative intersection over the cumulative union. Unless specified, we use gIoU as our default metric, as it equally considers both large and small objects.

% \begin{table}[t]
%   \centering
%   \scalebox{1.0}{
%       \begin{tabular}{llccc}
%         \toprule
%         Model & Type  & CoT & TextVQA & ChartQA  \\
%         \midrule
%         Baseline & & & 92.8 & 83.6  \\
%         Seg-Zero & SFT  & \texttimes & 81.9 & 75.2 \\ % r1v id 0
%         Seg-Zero & RL & \texttimes & 92.6 & 83.6 \\ % r1v id 11
%         Seg-Zero & RL & \checkmark & \textbf{93.1} & \textbf{83.8} \\ % id = 10
%         \bottomrule
%       \end{tabular}
%   }
%   \caption{Performance comparison on Visual QA tasks.}
%   \label{tab:sft_vs_rl_vqa}
% \end{table}

\subsection{SFT vs.~RL}
\begin{figure}[t]
  \centering
   \includegraphics[width=1.0\linewidth]{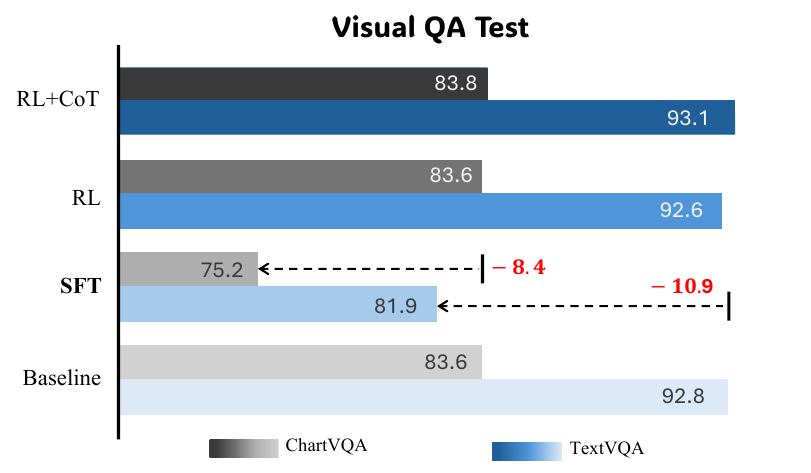}
   \caption{Visual QA task comparison. SFT suffers catastrophic forgetting, while RL preserves general Visual QA ability.}
   \label{fig:sft_vs_rl_vqa}
\end{figure}

We compare the performance of SFT and RL. The baseline model is Qwen2.5-VL-3B + SAM2-Large. 
%In this part, we train the SFT model on the entire RefCOCOg training set, select the best-performing checkpoint, whereas train the RL model only on 9,000 samples. 
For the non-CoT setting, we eliminate the thinking format reward, thus the model does not generate a CoT reasoning process before outputting the final answer. Our comparison includes both in-domain and OOD segmentation tasks \cite{singh2019textvqa,masry2022chartqa}, as well as general QA tasks. The corresponding results are shown in \Cref{tab:sft_vs_rl}, \Cref{fig:teaser_figure} and \Cref{fig:sft_vs_rl_vqa}.

\setlength{\tabcolsep}{5pt}{
\begin{table}[t]
    \footnotesize
    \centering
    \scalebox{1.0}{
        \begin{tabular}{llccc}
            \toprule
            Model & Type  & CoT & RefCOCOg & ReasonSeg  \\
            \midrule
            Baseline & & & 70.4 & 47.6 \\
            Seg-Zero & SFT  & \texttimes & 70.8 & 44.9  \\
            Seg-Zero & RL & \texttimes & 73.2 & 51.3  \\ % id = r1v some
            Seg-Zero & RL & \checkmark & \textbf{73.6} & \textbf{53.8}  \\ % id = 10
            \bottomrule
        \end{tabular}
    }
    \caption{Segmentation task comparison. Model trained with RL + CoT thinking reward achieves best performance on in-domain and OOD data.}
    \label{tab:sft_vs_rl}
\end{table}}

\minisection{SFT vs.~RL without CoT}
From the first two rows in \Cref{tab:sft_vs_rl}, we observe that on the in-domain dataset RefCOCOg, SFT achieves nearly the same performance as the baseline model. This may be due to the strong baseline performance of the original Qwen2.5-VL-3B. However, its performance significantly declines on the OOD ReasonSeg dataset, suggesting that SFT negatively impacts the model's generalization ability. In contrast, comparing the first and third rows, we find that RL consistently improves performance on both in-domain and OOD datasets, demonstrating the effectiveness of RL. Besides, from \Cref{fig:sft_vs_rl_vqa}, we observe that the SFT model suffers from catastrophic forgetting of its original visual QA ability, while the RL model effectively preserves this capability.

\minisection{RL without CoT vs.~RL with CoT}
From the last two rows in \Cref{tab:sft_vs_rl}, we find that both RL and RL with CoT achieve superior performance on both the in-domain RefCOCOg and OOD ReasonSeg datasets, significantly outperforming the baseline. This indicates that RL effectively boosts the models' capabilities. However, with CoT, our Seg-Zero demonstrates even better performance compared to its counterparts without CoT, indicating that the reasoning process enhances the model's ability to handle OOD data samples. From \Cref{fig:sft_vs_rl_vqa}, it is noteworthy that the introduction of CoT reasoning leads to a slight performance improvement in visual QA tasks for models trained without CoT.

% Recent works compared CoT RL with 
% reason seg (OOD)
% No harm to original models' output (keep QA ability)
\subsection{Ablation Study}
We conduct several ablation studies to verify the effectiveness of our design. For the ablation study, the default settings are as follows: we perform reinforcement learning using the GRPO algorithm on 9,000 samples and evaluate the model on the RefCOCOg test and the ReasonSeg test.

\setlength{\tabcolsep}{6pt}{
\begin{table}[t]
    \footnotesize
  \centering
  \scalebox{1.0}{
      \begin{tabular}{lcccc}
        \toprule
        Model & Bbox & Points  & RefCOCOg & ReasonSeg  \\
        \midrule
        Baseline &  & & 70.4 & 47.6 \\
        Seg-Zero & \texttimes  & \checkmark & 69.6 &  45.5 \\
        Seg-Zero & \checkmark & \texttimes & 72.9  & 53.6 \\
        Seg-Zero & \checkmark & \checkmark & \textbf{73.6} & \textbf{53.8}  \\ % id=10
        \bottomrule
      \end{tabular}
  }
  \caption{Ablation on the design of bbox and points prompt.}
  \label{tab:diff_prompts}
\end{table}}

\minisection{Design of Bbox and Points}
% The difference between our Seg-Zero and the baseline Qwen2.5-VL-3B is the addition of points prompt. 
\Cref{tab:diff_prompts} demonstrates the effectiveness of our bbox and points prompt design. We observe that using only point prompts results in worst performance. When both bbox and point prompts are utilized, Seg-Zero achieves its best performance, indicating that the combination of these prompts enhances pixel-level localization accuracy.

\setlength{\tabcolsep}{10.5pt}{
\begin{table}[t]
    \footnotesize
    \centering
    \scalebox{1.0}{
        \begin{tabular}{lcccc}
            \toprule
            Model & coef   & RefCOCOg & ReasonSeg  \\
            \midrule
            Seg-Zero & 1e-3 & \textbf{73.7} & 52.7  \\ % run id 9
            Seg-Zero & 5e-3 & 73.6 & \textbf{53.8}  \\ % run id 10
            Seg-Zero & 1e-2 & 70.6 & 53.3 \\ % run id 11
            Seg-Zero & 5e-2 & 66.6 & 50.8 \\ % run id 8
            \bottomrule
        \end{tabular}
    }
    \caption{
        Ablation on the KL loss coefficient.
        This coefficient balance `pre-existing knowledge' and `new knowledge'.
        Higher coefficient causes performance degradation.
    }
    \label{tab:diff_kl_coef}
\end{table}}

\minisection{KL Loss Coefficient}
The KL loss coefficient balances the model's `pre-existing knowledge' with `new knowledge'. \Cref{tab:diff_kl_coef} presents the performance variations across different KL loss coefficients. We find that a coefficient of 5e-3 performs optimally on both in-domain and OOD data. A higher coefficient leads to performance degradation. 
%Through qualitative analysis, we observe that when the coefficient is set to 1e-3, the model tends to output the user prompt `\textit{\texttt{<think>} thinking process here \texttt{<\textbackslash think>}}' instead of an actual reasoning process.

\setlength{\tabcolsep}{9pt}{
\begin{table}[t]
    \footnotesize
  \centering
  \scalebox{1.0}{
      \begin{tabular}{lcccc}
        \toprule
        Model & Number  & RefCOCOg & ReasonSeg  \\
        \midrule
        Seg-Zero & 4 & 66.8 & 52.0 \\ % run id 13
        Seg-Zero & 8 & 73.6  & 53.8  \\ % run id 10
        Seg-Zero & 16 & \textbf{74.1} & \textbf{54.7} \\ % run id 14
        \bottomrule
      \end{tabular}
  }
  \caption{Ablation on number of samples. A larger sample number leads to better performance.}
  \label{tab:num_samples}
\end{table}}

\minisection{Number of Samples}
We investigate the impact of the number of samples during the sampling stage. As shown in \Cref{tab:num_samples}, we observe that as the number of samples increases, the model achieves better performance on both in-domain and out-of-distribution (OOD) data. This is reasonable because a larger number of samples expands the exploration space, enabling the model to identify more effective optimization directions.

\setlength{\tabcolsep}{8pt}{
\begin{table}[t]
    \footnotesize
    \centering
    \scalebox{1.0}{
        \begin{tabular}{lcccc}
            \toprule
            Model & Example   & RefCOCOg & ReasonSeg  \\
            \midrule
            Seg-Zero & \texttimes & 72.1  & 49.4 \\ % run id 16
            Seg-Zero & \checkmark & \textbf{73.6} & \textbf{53.8}  \\ % run id 10
            \bottomrule
        \end{tabular}
    }
    \caption{Ablation on the the user prompt. Including an example in the user prompt is crucial.}
    \label{tab:user_prompt}
\end{table}}

\minisection{User Prompt Sensitivity}
The last two rows of \Cref{tab:user_prompt_template} show that we include output examples in the user prompt.
We investigate the impact of this example in \Cref{tab:user_prompt} and observe that its inclusion significantly enhances the model's performance. Through analysis of the output, we find that models without this example often fail to generate a reasoning process in their responses.

\setlength{\tabcolsep}{6pt}{
\begin{table}[t]
    \footnotesize
  \centering
  \scalebox{1.0}{
      \begin{tabular}{llcc|c}
        \toprule
        Model & Type   & RefCOCOg & ReasonSeg & sum \\
        \midrule
        Seg-Zero & Soft & 70.2  & \textbf{54.1} & 124.3 \\ % run id 12
        Seg-Zero & Hard & \textbf{73.6} & 53.8 & \textbf{127.4} \\ % run id 10
        \bottomrule
      \end{tabular}
  }
  \caption{Ablation on the accuracy reward type.}
  \label{tab:acc_reward_type}
\end{table}}

\minisection{Soft vs.~Hard Accuracy Rewards}
In \Cref{sec:rewards}, we describe the bbox IoU reward, the bbox L1 reward, and the point L1 reward.
We apply specific thresholds to convert these metrics into binary rewards.
Additionally, we conduct ablation studies on soft counterparts.
For the bbox IoU reward, we directly use the IoU value as the soft reward.
For L1-based rewards, we define the soft reward as $1-\dfrac{\text{L1 dist}}{\max\{\text{image size}\}}$.
From \Cref{tab:acc_reward_type}, we observe that while the soft reward achieves a minor improvement on ReasonSeg,
it significantly underperforms compared to the hard reward on RefCOCOg.

\setlength{\tabcolsep}{6pt}{
\begin{table}[t]
    \footnotesize
  \centering
  \scalebox{1.0}{
      \begin{tabular}{llcc|c}
        \toprule
        Model & Type   & RefCOCOg & ReasonSeg & sum  \\
        \midrule
        Seg-Zero & Soft & \textbf{73.6} & 53.8 & 127.4 \\ % run id 10
        Seg-Zero & Strict & 73.0 & \textbf{56.1} & \textbf{129.1}  \\ % run id 15
        \bottomrule
      \end{tabular}
  }
  \caption{Ablation on the format reward type. Strict format is better.}
  \label{tab:format_reward_type}
\end{table}}

\minisection{Soft vs.~Strict Format Rewards}
In \Cref{sec:rewards}, we introduce two types of segmentation format rewards: the soft and strict.
From \Cref{tab:format_reward_type}, we find that the strict format reward significantly improves performance gain on OOD data in ReasonSeg. Through qualitative analysis of the training steps, we find that the strict format reward progresses slowly in the initial stages, as it is more challenging to sample formats that precisely match the strict criteria. However, as training step increases, model with strict format reward tend to output longer response. % \textcolor{red}{the response length comparison here.}

\setlength{\tabcolsep}{11pt}{
\begin{table}[t]
    \footnotesize
  \centering
  \scalebox{1.0}{
      \begin{tabular}{lcc}
        \toprule
        Reasoning Model   & RefCOCOg &  ReasonSeg  \\
        \midrule
        Qwen2-VL-2B   &   70.1 & 37.2 \\ % id = 23
        % Qwen2.5-VL-3B &  SAM &  &   \\
        Qwen2.5-VL-3B &   73.0  & 56.1 \\ % id = 15
        % Qwen2.5-VL-3B &  TAM & &   \\
        Qwen2.5-VL-7B &  \textbf{74.2}  & \textbf{57.5}  \\ % id = 22
        \bottomrule
      \end{tabular}
  }
  \caption{Ablation on reasoning model choice. Larger scale model achieves better performance.}
  \label{tab:diff_reasoning_model}
\end{table}}

\begin{figure}[t]
  \centering
   \includegraphics[width=0.86\linewidth]{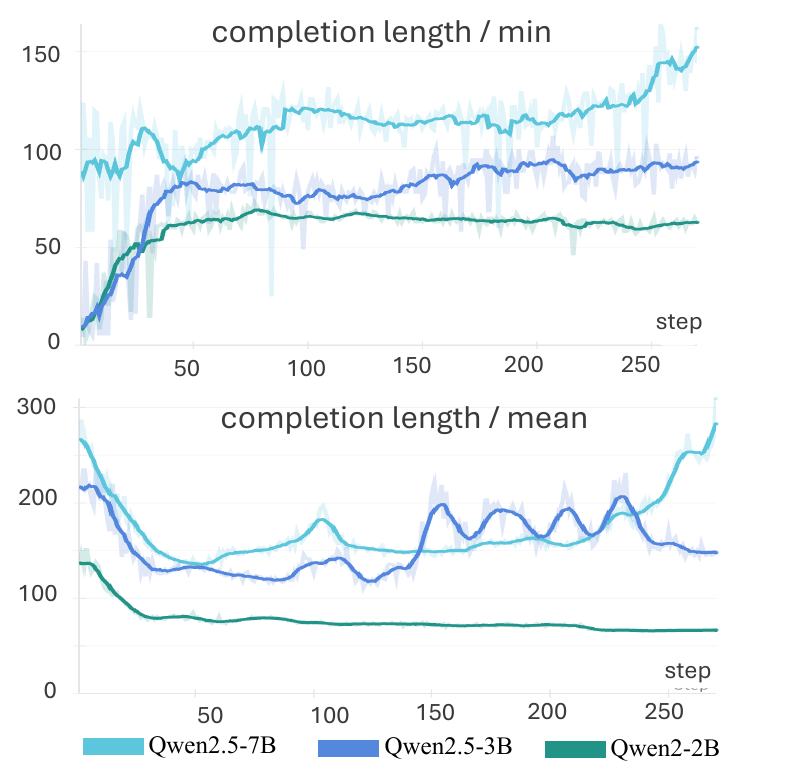}
   \caption{Changes in completion length during training. Larger scale model tends to generate longer response.}
   \label{fig:response_length}
\end{figure}

\minisection{Reasoning Model Scale}
We conduct an ablation study on reasoning models of varying scales, ranging from 2B to 7B parameters, under the same rewards and training settings. As shown in \Cref{tab:diff_reasoning_model}, we observe that model performance on both in-domain and OOD data improves as the model scale increases. 

\minisection{Changes in Completion Length}
\Cref{fig:response_length} illustrates the trends in completion lengths between different model sizes. The results indicate that a larger model tends to generate longer responses. 
As training progresses, the minimal completion length gradually increases.
However, there is a drop in average completion length during the initial few steps.
By analyzing the output during the training process, we find that this occurs because the model initially prioritizes learning the correct output format, which often results in shorter responses.
Once the format reward saturates, the model shifts its focus to generating answer with higher accuracy, leading to longer and more detailed responses.
\textit{Supplementary materials} provide more analysis.

\setlength{\tabcolsep}{6pt}{
\begin{table}[t]
    \footnotesize
\centering
\begin{tabular}{l|cc|cc}
\toprule
\multirow{3}{*}{Method} & \multicolumn{4}{c}{ReasonSeg} \\
\cline{2-5}
 & \multicolumn{2}{c|}{val} & \multicolumn{2}{c}{test} \\
\cline{2-5}
 & gIoU & cIoU &  gIoU & cIoU \\
\midrule
OVSeg      & 28.5 & 18.6 & 26.1 & 20.8  \\
ReLA       & 22.4 & 19.9 & 21.3 & 22.0  \\
% X-Decoder  & 22.6 & 17.9 & 21.7 & 16.3  \\
% SEEM       & 25.5 & 21.2 & 24.3 & 18.7  \\
Grounded-SAM & 26.0 & 14.5 & 21.3 & 16.4 \\
% LISA-7B    & 44.4 & 46.0 & 36.8 & 34.1 \\
% LISA-13B   & 48.9 & 46.9 & 44.8 & 45.8 \\
LISA-7B-LLaVA1.5 & 53.6 & 52.3 & 48.7 & 48.8 \\
LISA-13B-LLaVA1.5 & 57.7 & 60.3 & 53.8 & 50.8 \\
SAM4MLLM   & 46.7 & 48.1 & - & - \\
Qwen2.5VL-3B+SAM2 & 53.8 & 44.1 & 47.6 & 37.4 \\
\cellcolor[HTML]{efefef}{Seg-Zero-3B (ours)} & \cellcolor[HTML]{efefef}{58.2} & \cellcolor[HTML]{efefef}{53.1} & \cellcolor[HTML]{efefef}{56.1} & \cellcolor[HTML]{efefef}{48.6} \\ % id=15
\cellcolor[HTML]{efefef}{Seg-Zero-7B (ours)} & \cellcolor[HTML]{efefef}{\textbf{62.6}} & \cellcolor[HTML]{efefef}{\textbf{62.0}} & \cellcolor[HTML]{efefef}{\textbf{57.5}} & \cellcolor[HTML]{efefef}{\textbf{52.0}} \\ % id = 22 

\bottomrule
\end{tabular}
\caption{Zero-shot reasoning segmentation results.}
\label{table:performance_comparison_reasoning}
\end{table}}

\setlength{\tabcolsep}{4pt}{
\begin{table}[t]
    \footnotesize
\centering
\begin{tabular}{l|ccc}
\toprule
Method & \multicolumn{1}{c}{RefCOCO} & \multicolumn{1}{c}{RefCOCO+} & \multicolumn{1}{c}{RefCOCOg} \\
                  & testA & testA  &  test \\
\midrule
%MCN         & 62.4 & 64.2 & 59.7 & 50.6 & 55.0 & 44.7 & 49.2 & 49.4 \\
%VLT         & 67.5 & 70.5 & 65.2 & 56.3 & 61.0 & 50.1 & 55.0 & 57.7 \\
%CRIS        & 70.5 & 73.2 & 66.1 & 62.3 & 68.1 & 53.7 & 59.9 & 60.4 \\
LAVT               & 75.8  & 68.4 &  62.1 \\
ReLA               & 76.5  & 71.0 &  66.0 \\
LISA-7B            & 76.5  & 67.4 &  68.5 \\
PixelLM-7B         & 76.5  & 71.7 &  70.5 \\
PerceptionGPT-7B   & 78.6  & 73.9 &  71.7 \\
% SAM4MLLM-7B        & 82.8  & 77.8 &  75.6 \\
% Qwen2.5VL-3B+SAM2 & \\
\cellcolor[HTML]{efefef}{Seg-Zero-3B (ours)} & \cellcolor[HTML]{efefef}{79.3}  &  \cellcolor[HTML]{efefef}{73.7}& \cellcolor[HTML]{efefef}{71.5}\\ % id=14
\cellcolor[HTML]{efefef}{Seg-Zero-7B (ours)} & \cellcolor[HTML]{efefef}{\textbf{80.3}}  & \cellcolor[HTML]{efefef}{\textbf{76.2}} &  \cellcolor[HTML]{efefef}{\textbf{72.6}}\\ % id = 17
\bottomrule
\end{tabular}
\caption{Referring expression segmentation results. We compare cIoU in this table.}
\label{tab:performance_comparison_referring}
\end{table}}

\begin{figure*}[tb!]
  \centering
   \includegraphics[width=1.0\linewidth]{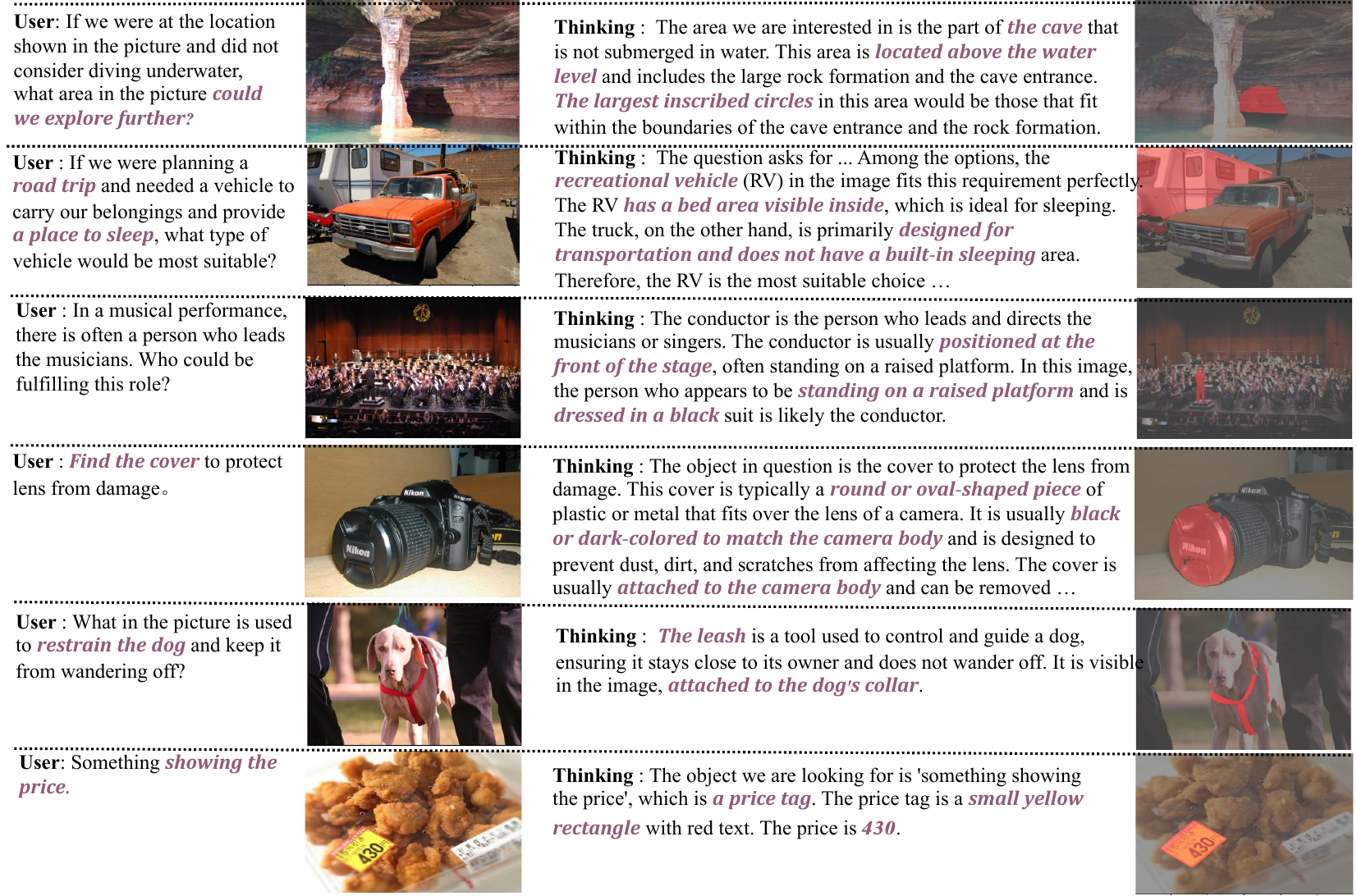}
   \caption{Qualitative Results on ReasonSeg \cite{lai2024lisa}. The reason chain helps analyze user instructions and segment the correct objects.}
   \label{fig:qualitative}
\end{figure*}

\subsection{Comparison with Other Methods}

%We compare the performance on referring expression segmentation tasks and reasoning segmentation tasks. 
In this part, we train our Seg-Zero using hard accuracy rewards and strict format rewards. The sampling number is set to 16. And we only train our Seg-Zero on 9,000 samples from RefCOCOg. We compare OVSeg \cite{liang2023ovseg}, Grounded-SAM \cite{ren2024groundedsam}, LISA \cite{lai2024lisa}, SAM4MLLM \cite{chen2024sam4mllm}, LAVT \cite{yang2022lavt}, ReLA \cite{liu2023rela}, PixelLM \cite{ren2024pixellm}, PerceptionGPT \cite{pi2024perceptiongpt}. 

\minisection{Reasoning Segmentation}
We compare the zero-shot performance on ReasonSeg \cite{lai2024lisa}, results are shown in \Cref{table:performance_comparison_reasoning}. We can find our Seg-Zero achieves the SOTA \textcolor{mycolor}{zero}-shot performance across various methods. 

\minisection{Referring Expression Segmentation}
The results on referring expression segmentation are shown on \Cref{tab:performance_comparison_referring}. Moreover, we find that the ground-truth annotations in RefCOCO(+/g) are not precise enough, which suggests that our Seg-Zero model should, in principle, achieve better performance than values in the table. \textit{Supplementary materials} provide detailed analysis.

% (todo) reasonseg
% refcoco g/+

% zero-shot OV-Seg
% zero-shot Object detect

\subsection{Qualitative Results}

We provide several examples in \Cref{fig:qualitative}.
We can easily observe that the reasoning process is helpful in analyzing user instructions, especially when there are multiple objects within the same class categories.
For instance, Seg-Zero demonstrates its ability to discern that a `recreational vehicle' is more appropriate than a `truck' in the context of a `road trip',
and correctly identifies that a `conductor' is `positioned at the front of the stage'. 

% \subsection{Auxilary Reasoning Grounding Performance} 

\section{Conclusion}

% \textcolor{mycolor}{Zero}
In this paper, we propose Seg-\textcolor{mycolor}{Zero}, a novel framework that integrates the CoT reasoning process into segmentation tasks. We design a sophisticated reward mechanism, incorporating both format and accuracy constraints, to guide the optimization directions. By training exclusively with RL, Seg-Zero emerges reasoning capabilities without relying on any supervised reasoning data. We present a detailed comparison between SFT and RL, as well as the introduction of reason chain. Additionally, we offer insightful perspectives on the design of RL and the reward functions.

\clearpage
\appendix
\section{Rewards Changes During Training}
\begin{figure}[t]
  \centering
   \includegraphics[width=0.8\linewidth]{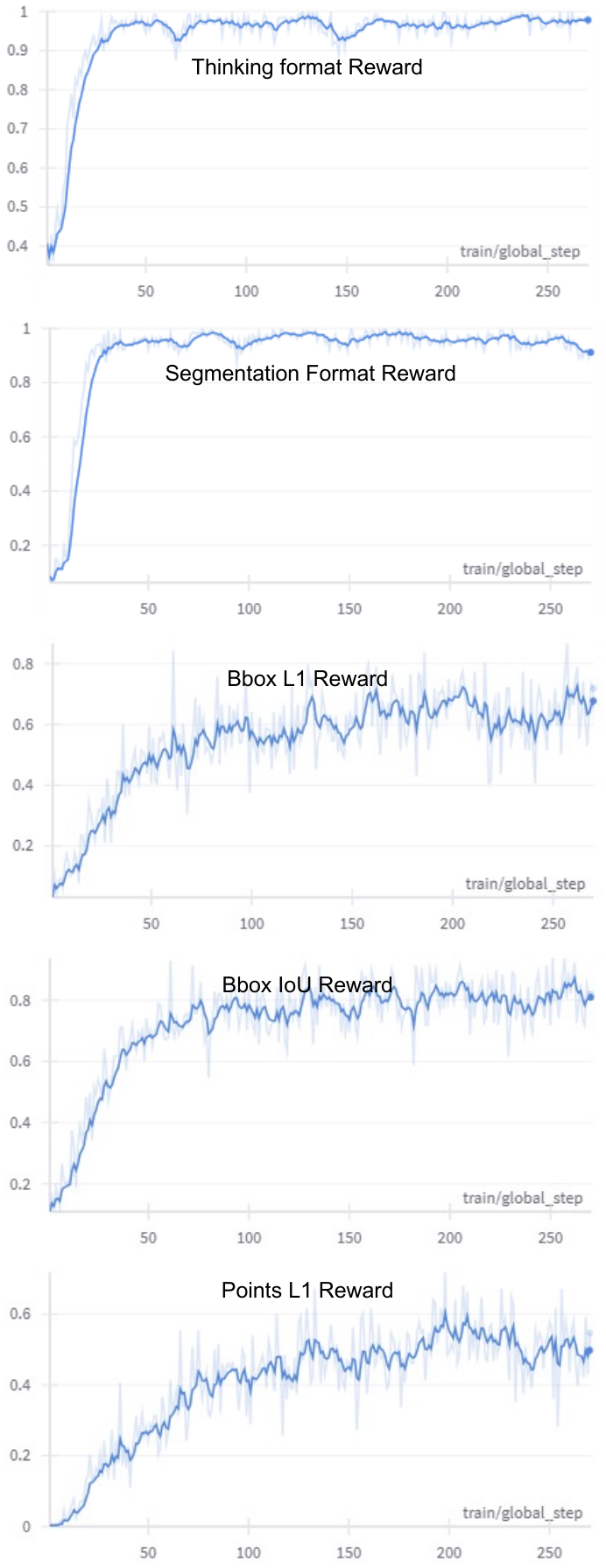}
   \caption{Changes in rewards during training. We show the mean value across a batch.}
   \label{fig:supp_rewards}
\end{figure}

We visualize the changes in rewards during the training process. As shown in \Cref{fig:supp_rewards}, the format rewards converge to 1 in few steps, and the accuracy rewards gradually increase over time. This suggests that the format rewards initially dominates the optimization direction, leading to a decrease in response length during the initial training steps, as shown in \Cref{fig:response_length}. However, as the format rewards converge and the accuracy rewards gradually increases, the model's completion length (i.e., the CoT reasoning process) begins to expand.

\section{More Visualization Examples}
\begin{figure*}[t]
  \centering
   \includegraphics[width=0.9\linewidth]{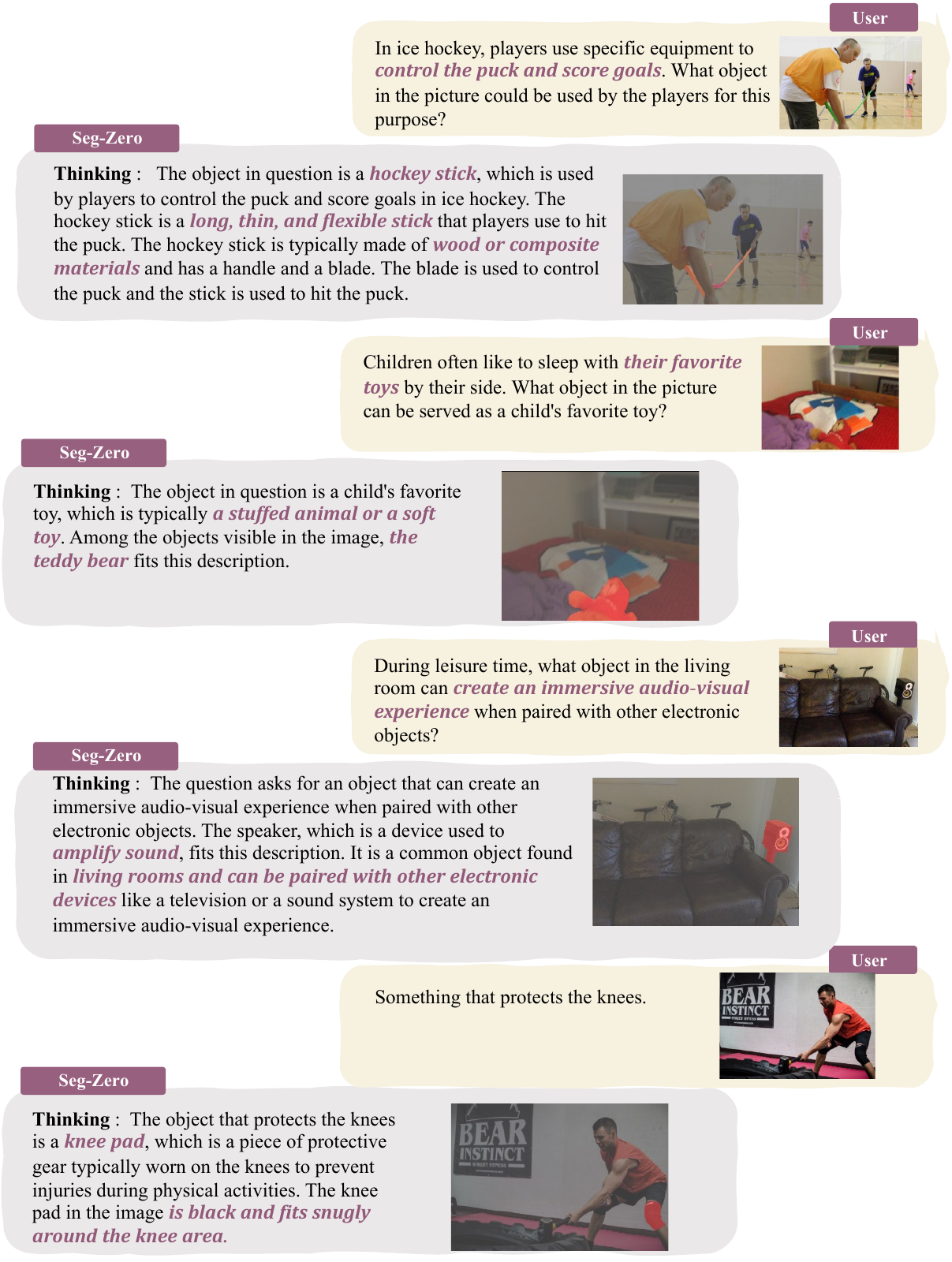}
   \caption{More examples.}
   \label{fig:supp_examples}
\end{figure*}

We provide more examples in \Cref{fig:supp_examples}.

\section{Additional Analysis of RefCOCO(+/g)}

\begin{figure*}[t]
  \centering
   \includegraphics[width=0.9\linewidth]{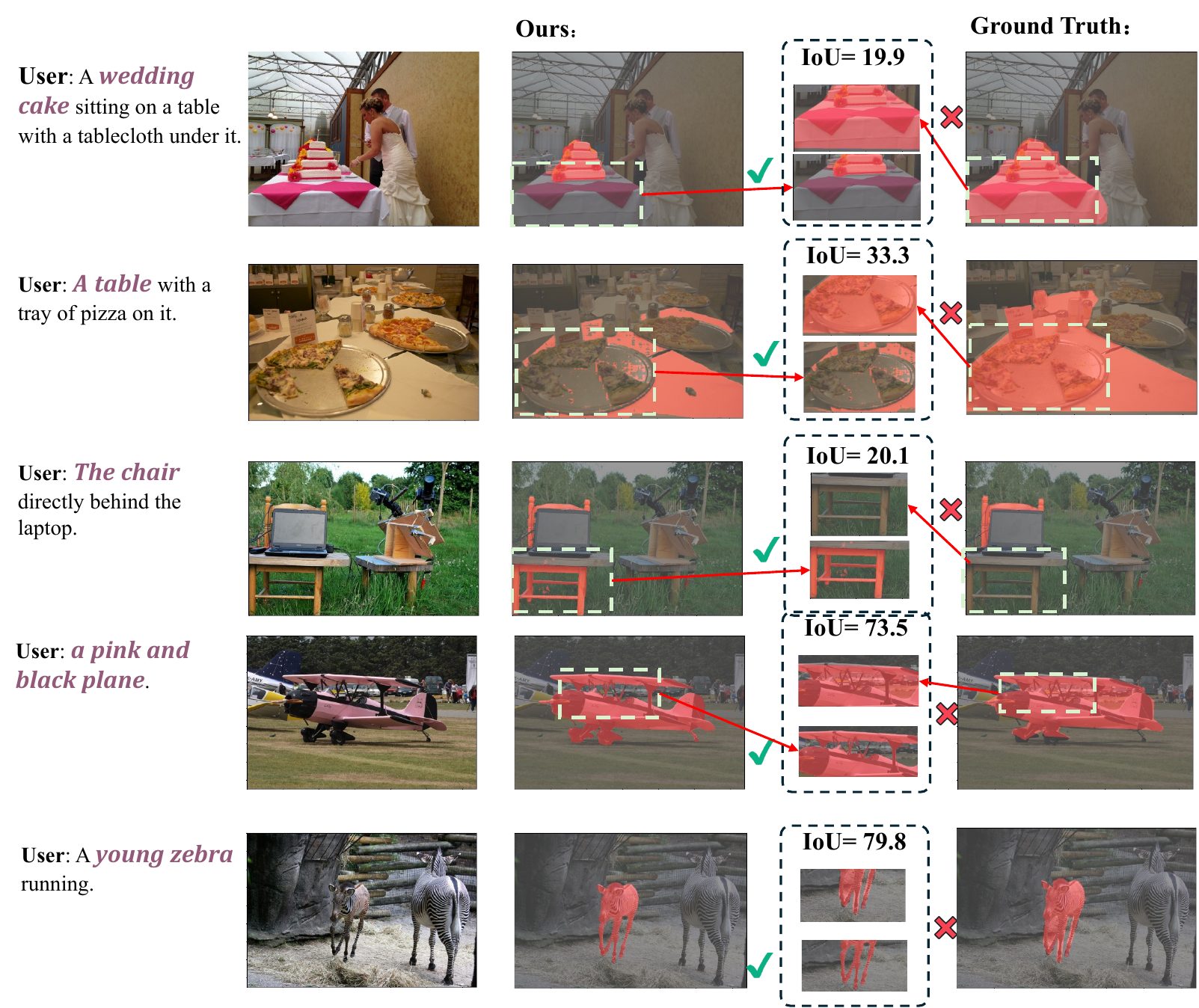}
   \caption{Visual analysis of RefCOCO(+/g). The grouth-truth label on RefCOCO(+/g) is not precious, whereas our results are precious.}
   \label{fig:supp_refcoco}
\end{figure*}

Through an analysis of the RefCOCO(+/g) datasets, we observe that the mask annotations in these datasets lack precision. Specifically, these annotations either inaccurately represent object sizes or exhibit imprecise edge handling, as illustrated in \Cref{fig:supp_refcoco}. We randomly select 100 samples, and find that 5\% samples has `inaccurately represent object sizes ', and almost 70\% samples is not precious in `edge hanlding'. Although our model demonstrates high precious, these annotation errors result in a lower IoU score.
% This issue has also been highlighted by SAM4MLLM \cite{chen2024sam4mllm}. 

Since previous methods \cite{lai2024lisa,chen2024sam4mllm} fine-tunes the segmentation decoder on these datasets, they can adapt to these imprecise annotations. In contrast, our segmentation model remains frozen and maintains higher precision. Moreover, since the gIoU computes the aggregate of this bias, it fails to accurately reflect the true performance of our model. Therefore, the actual precious of our Seg-Zero should be better than the performance presented in \Cref{tab:performance_comparison_referring}.
To further evaluate the localization accuracy of our Seg-Zero model, we introduce two additional metrics: bbox accuracy and point accuracy. The bbox accuracy metric computes whether the IoU between the predicted bounding box and the maximum bounding box that fully encloses the ground-truth mask exceeds 0.5. Meanwhile, the point accuracy metric assesses whether the predicted points lie within the ground-truth mask. These metrics provide insights into the model's ability to precisely localize the target object.
As shown in \Cref{table:supp_bbox_point_acc}, our Seg-Zero model accurately localizes the target object with a success rate of approximately 90\%.

\setlength{\tabcolsep}{10pt}{
\begin{table}[t]
\centering
\begin{tabular}{l|cc|cc}
\hline
\multirow{3}{*}{Method} & \multicolumn{4}{c}{ReasonSeg} \\
\cline{2-5}
 & \multicolumn{2}{c|}{val} & \multicolumn{2}{c}{test} \\
\cline{2-5}
 & bbox & point & bbox & point \\
\hline
\cellcolor[HTML]{efefef}{Seg-Zero-3B} & \cellcolor[HTML]{efefef}{85.4} & \cellcolor[HTML]{efefef}{89.3} & \cellcolor[HTML]{efefef}{85.5} & \cellcolor[HTML]{efefef}{88.8} \\ % id=14
\cellcolor[HTML]{efefef}{Seg-Zero-7B} & \cellcolor[HTML]{efefef}{\textbf{85.4}} & \cellcolor[HTML]{efefef}{\textbf{89.0}} & \cellcolor[HTML]{efefef}{\textbf{86.6}} & \cellcolor[HTML]{efefef}{\textbf{90.0}} \\ % id = 17

\hline
\end{tabular}
\caption{Bbox IoU and Point Accuracy.}
\label{table:supp_bbox_point_acc}
\end{table}}

\section{Limitations and Future Works}
Currently, our Seg-Zero only focuses on single object reasoning segmentation. We are considering employ our method to multi-object reasoning segmentation, instance-level reasoning segmentation.

{
    \small
    \bibliographystyle{ieeenat_fullname}
    \bibliography{main}
}

\end{document}